\documentclass[letterpaper, 10 pt, conference]{ieeeconf}  

\IEEEoverridecommandlockouts                              

\usepackage{cite}
\usepackage{amsmath,amssymb,amsfonts}
\usepackage{graphicx}
\usepackage{cuted}
\usepackage{caption}
\usepackage{textcomp}
\usepackage{xcolor}
\def\BibTeX{{\rm B\kern-.05em{\sc i\kern-.025em b}\kern-.08em
    T\kern-.1667em\lower.7ex\hbox{E}\kern-.125emX}}

\usepackage{mathrsfs}
\usepackage{mathtools}
\usepackage{pbox}
\usepackage{xcolor,colortbl}

\usepackage{wrapfig}
\usepackage{physics}
\usepackage{graphicx}
\usepackage{hyperref}

\usepackage{pgfplots}
\usepgfplotslibrary{groupplots,dateplot}
\usetikzlibrary{patterns,shapes.arrows}
\pgfplotsset{compat=newest}
\usepackage{tikz}
\usetikzlibrary{shapes, arrows.meta, positioning}
\usetikzlibrary{decorations.pathreplacing,calligraphy}

\usepgfplotslibrary{groupplots}
\usepgfplotslibrary{fillbetween}
\usetikzlibrary{arrows,decorations.pathmorphing,positioning,fit,trees,shapes,shadows,automata,calc} 
\usetikzlibrary{3d,patterns,arrows,arrows.meta,calc,shapes,shadows,decorations.pathmorphing,decorations.pathreplacing,automata,shapes.multipart,positioning,shapes.geometric,fit,circuits,trees,shapes.gates.logic.US,fit, matrix,arrows.meta, quotes}
\usetikzlibrary{backgrounds,scopes}
\usepackage{neuralnetwork}
\usepackage{listofitems} 

\usepgfplotslibrary{external}
\pgfplotsset{compat=newest}
\newenvironment{customlegend}[1][]{%
	\begingroup
	\csname pgfplots@init@cleared@structures\endcsname
	\pgfplotsset{#1}%
}{%
	\csname pgfplots@createlegend\endcsname
	\endgroup
}%

\def\addlegendimage{\csname pgfplots@addlegendimage\endcsname}
\pgfdeclarelayer{background}
\pgfdeclarelayer{background2}
\pgfdeclarelayer{background1}
\pgfdeclarelayer{foreground1}
\pgfdeclarelayer{foreground2}
\pgfsetlayers{background,background2,background1,main,foreground1,foreground2}

\begin{document}

\title{\LARGE \bf
A Multifidelity Sim-to-Real Pipeline for \\ Verifiable and Compositional Reinforcement Learning
}

\author{Cyrus Neary$^{1}$, Christian Ellis$^{2}$, Aryaman Singh Samyal$^{1}$, Craig Lennon$^{3}$, and Ufuk Topcu$^{1}$
\thanks{$^{1}$The University of Texas at Austin
        {\tt\small \{cneary, aryamansinghsamyal, utopcu\}@utexas.edu}}%
\thanks{$^{2}$The University of Massachusetts Dartmouth
        {\tt\small cellis3@umassd.edu}}%
\thanks{$^{3}$U.S. Army Research Lab
        {\tt\small craig.t.lennon.civ@army.mil}}
}

\maketitle

\newcommand{\defeq}{\vcentcolon=}

\definecolor{planner}{HTML}{E05F15}
\definecolor{rlPolicies}{HTML}{07742D}
\definecolor{teleOp}{HTML}{4F359B}
\definecolor{rlPoliciesHighSim}{HTML}{00A1E0}
\definecolor{rlPoliciesLowSim}{HTML}{130303}

\definecolor{waypoint}{HTML}{ffd966}
\definecolor{goal}{HTML}{03C91D}

\newcommand{\waypointOpacity}{0.8}
\newcommand{\waypointSize}{3}

\newcommand{\trajectoryWidth}{1}

\newcommand{\colorpar}[3]{\colorbox{#1}{\parbox{#2}{#3}}}
\newcommand{\marginremark}[3]{\marginpar{\colorpar{#2}{\linewidth}{\color{#1}#3}}}
\newcommand{\cn}[1]{\marginremark{red}{white}{\scriptsize{[CN]~ #1}}}
\newcommand{\mc}[1]{\marginremark{red}{white}{\scriptsize{[MC]~ #1}}}

\newcommand{\bernoulliProb}{p}
\newcommand{\rewardRV}{R}
\newcommand{\rewardFunction}{R}
\newcommand{\rewardExpectedVal}{r}

\newcommand{\mdp}{M}
\newcommand{\mdpStateSet}{S}
\newcommand{\mdpState}{s}
\newcommand{\mdpActionSet}{A}
\newcommand{\mdpAction}{a}
\newcommand{\mdpRewardFunction}{R}
\newcommand{\mdpCommonReward}{r}
\newcommand{\mdpTransition}{P}
\newcommand{\mdpDiscount}{\gamma}
\newcommand{\mdpInitialState}{\mdpState_I}
\newcommand{\mdpInitialDist}{\alpha_I}
\newcommand{\mdpCostFunction}{C}

\newcommand{\pomdp}{\mdp}
\newcommand{\pomdpObservationSet}{Z}
\newcommand{\pomdpObservation}{z}
\newcommand{\pomdpObservationFunction}{\mathcal{O}}

\newcommand{\distribution}{\Delta}

\newcommand{\stateAbstraction}{\phi}

\newcommand{\policy}{\pi}
\newcommand{\targetStateSet}{\mdpStateSet_{targ}}
\newcommand{\initialStateSet}{\mdpStateSet_{init}}
\newcommand{\timeHorizon}{T}
\newcommand{\mdpSuccessState}{\mdpState_{g}}
\newcommand{\valueFunction}{V}
\newcommand{\terminationTime}{\tau}

\newcommand{\abstractMDP}{\tilde{\mdp}}
\newcommand{\abstractStateSet}{\tilde{\mdpStateSet}}
\newcommand{\abstractState}{\tilde{\mdpState}}
\newcommand{\abstractActionSet}{\tilde{\mdpActionSet}}
\newcommand{\abstractAction}{\tilde{\mdpAction}}
\newcommand{\abstractTransition}{\tilde{\mdpTransition}}
\newcommand{\abstractRewardFunction}{\tilde{\mdpRewardFunction}}
\newcommand{\abstractFailureState}{\abstractState_{\times}}
\newcommand{\abstractSuccessState}{\abstractState_{g}}
\newcommand{\abstractPolicy}{\mu}
\newcommand{\abstractInitialState}{\tilde{\mdpInitialState}}
\newcommand{\abstractInitialStateSet}{\abstractStateSet_{init}}
\newcommand{\abstractInitDist}{\alpha}
\newcommand{\abstractTargetStateSet}{\tilde{\targetStateSet}}

\newcommand{\boundMDP}{\bar{\mdp}}
\newcommand{\bernoulliProbBound}{\bar{\bernoulliProb}}
\newcommand{\boundMDPReward}{\bar{R}}
\newcommand{\boundMDPTransition}{\bar{\mdpTransition}}

\newcommand{\probThreshold}{P_{threshold}}
\newcommand{\failThreshold}{F_{threshold}}
\newcommand{\rewardThreshold}{R_{threshold}}

\newcommand{\hlmFailProb}{\delta}
\newcommand{\hlmPolicy}{\abstractPolicy}

\newcommand{\controller}{c}
\newcommand{\controllerSet}{\mathcal{C}}
\newcommand{\controllerInitialStateSet}{\mathcal{I}}
\newcommand{\controllerFinalStateSet}{\mathcal{F}}
\newcommand{\controllerTimeHorizon}{T}
\newcommand{\numControllers}{k}

\newcommand{\subtask}{\controller}
\newcommand{\subtaskSet}{\controllerSet}
\newcommand{\subtaskInitialStateSet}{\controllerInitialStateSet}
\newcommand{\subtaskFinalStateSet}{\controllerFinalStateSet}
\newcommand{\subtaskTimeHorizon}{\controllerTimeHorizon}
\newcommand{\numSubtasks}{\numControllers}

\newcommand{\successProb}{\sigma}

\newcommand{\eqRelation}{R}

\newcommand{\occupancyVar}{x}

\newcommand{\controllerInfProb}{\Bar{\successProb}}

\newcommand{\numSteps}{N}
\newcommand{\initTrainingSteps}{N_{init}}
\newcommand{\estimationRollouts}{N_{est}}
\newcommand{\trainingSteps}{N_{train}}
\newcommand{\maxTrainingSteps}{N_{max}}
\newcommand{\controllerPerformanceEstimate}{\hat{\successProb}}

\newcommand{\lbList}{\mathcal{L}}
\newcommand{\ubList}{\mathcal{U}}

\newcommand{\perfAwareOptProblem}{\Omega}

 \newcommand\titlesize{\fontsize{8.1pt}{10.2pt}\selectfont}

\newcommand{\rewardDistanceConstant}{C}

\newcommand{\entropy}{H}
\newcommand{\trajectory}{\tau}
\newcommand{\dataset}{\mathcal{D}}
\newcommand{\numTrajectories}{N}
\newcommand{\observedOccupancyVar}{\bar{\occupancyVar}^{\mdpDiscount}}

\newcommand{\hlmTimeHorizon}{N}
\newcommand{\metaDecisionTime}{\tau}
\newcommand{\numMetaDecision}{m}
\newcommand{\numTimeStep}{n}
\newcommand{\reachTrajectories}{\Gamma}
\newcommand{\sigmaAlg}{\Sigma}
\newcommand{\measure}{\mathbb{P}}
\newcommand{\history}{h}

\begin{figure*}[h]
    \centering
    \input{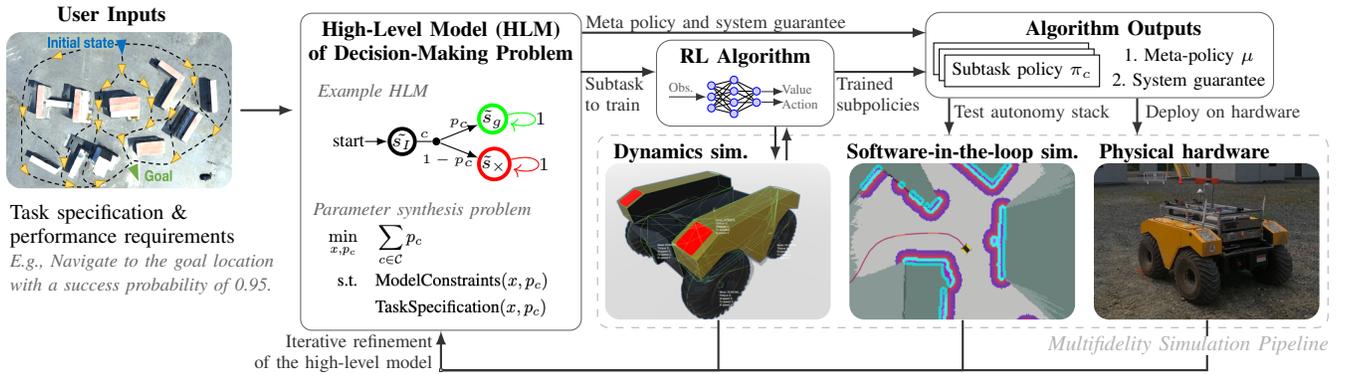}
    \caption{
    The proposed framework provides a compositional approach to training and verifying reinforcement learning (RL) policies in a multifidelity simulation pipeline, before deploying the trained policies on robot hardware.
    It constructs a \textit{high-level model} (HLM) of the robot's task, which it uses to: plan \textit{meta-policies} dictating which subtasks will be composed to complete the overall task, decompose task specifications into subtask specifications, select the subtasks whose RL-based policies require further training, assess the capabilities of the trained subtask policies, and use these assessments to place probabilistic guarantees on the success of the composite policy.
    A \textit{low-fidelity} simulation of the robot dynamics is initially used to train and empirically verify the subtask policies.
    A \textit{high-fidelity} software-in-the-loop simulation is used to test the integration of the trained policies with the existing autonomy software stack, before deploying the policies on the target robotic hardware.
    Feedback from both simulators and from the hardware tests is used to update the HLM, in order to adapt and re-train subtasks as necessary. 
    Videos of experiments are available at \url{https://tinyurl.com/44j8s8nz}.}
    \label{fig:abstract_figure}
\end{figure*}

\begin{abstract}
We propose and demonstrate a compositional framework for training and verifying reinforcement learning (RL) systems within a multifidelity sim-to-real pipeline, in order to deploy reliable and adaptable RL policies on physical hardware.
By decomposing complex robotic tasks into component subtasks and defining mathematical interfaces between them, the framework allows for the independent training and testing of the corresponding \textit{subtask policies}, while simultaneously providing guarantees on the overall behavior that results from their composition.
By verifying the performance of these subtask policies using a multifidelity simulation pipeline, the framework not only allows for efficient RL training, but also for a refinement of the subtasks and their interfaces in response to challenges arising from discrepancies between simulation and reality.
In an experimental case study we apply the framework to train and deploy a compositional RL system that successfully pilots a \textit{Warthog} unmanned ground robot.
\end{abstract}

\section{Introduction}
\label{sec:intro}

Recent progress in reinforcement learning (RL) shows tremendous promise in enabling data-driven robotic systems that learn to carry out complex tasks in a variety of operating environments \cite{mnih2015human,vinyals2019grandmaster,schrittwieser2020mastering,wurman2022outracing,degrave2022magnetic,songLimitsAutonomousRacing}.
However, the deployment of RL-trained policies on robotic hardware is challenging.
Training RL policies on robotic hardware can be costly, and simulations cannot perfectly capture the true system of interest\cite{balaji2019deepracer,degrave2022magnetic}.
This gap between simulation and reality may result in unwanted behaviors when simulation-based policies are deployed on hardware\cite{dulac2019challenges,zhao2020sim,ibarz2021train}.
Furthermore, it is challenging to verify the adherence of RL policies to user-defined specifications (\textit{e.g., complete the task of interest and avoid an unsafe set of states with a probability of at least \(0.95\)}).
However, a robot's ability to satisfy such specifications is often essential to its reliable and safe deployment \cite{baier2008principles,garcia2015comprehensive,amodei2016concrete,dulac2019challenges}.
These challenges become exacerbated when we consider complex tasks that require the execution of multiple subtasks over long time horizons.

We propose and demonstrate a compositional framework for training and verifying RL systems within a multifidelity sim-to-real pipeline, in order to deploy reliable and adaptable RL policies on physical hardware.
Figure \ref{fig:abstract_figure} illustrates the proposed framework, which consists of a multi-level abstraction of the decision-making problem itself, as well as the aforementioned multifidelity simulation pipeline.

Building on \cite{neary2022verifiable}, the framework defines a \textit{high-level model} (HLM), which is used to plan \textit{meta-policies} that select the subtasks necessary to complete the overall task.
The subtasks themselves are executed by \textit{subtask policies}, each of which is trained using an RL algorithm in simulation.
This multi-level approach might be viewed as a form of hierarchical RL \cite{sutton1999between,barto2003recent,kulkarni2016hierarchical,vezhnevets2017feudal,nachum2019data,levy2017learning,ji2022hierarchical}.
However, our objective is to formalize and take advantage of the task decompositions that are inherent to such algorithms in order to build a framework that facilitates sim-to-real transfer.

In particular, by formulating the HLM as a parametric Markov decision process (pMDP)~\cite{cubuktepe2018synthesis,junges2020parameter}, the framework enables automatic decomposition of task-level specifications into a collection of corresponding subtask specifications 
(\textit{e.g., reach the subtask's exit conditions with a probaility of 0.98 from its entry conditions})
via the solution to a parameter synthesis problem.
This decomposition allows for highly goal-oriented training and testing of the subtask policies.
It also enables an iterative procedure in which estimates of the capabilities of the subtask policies are used to update the HLM, which re-plans accordingly and selects the best subtasks to (re-)train and deploy.
Finally, it allows for efficient adaptation to changes in the decision-making problem: previously learned subtask policies can be reused as components of new HLMs that solve different tasks in altered environments.

The framework that we propose integrates a multifidelity sim-to-real pipeline with the iterative and compositional approach to RL described above.
In general, discrepancies between simulation and reality can result not only from a lack of accuracy in predicting the robot's physical dynamics, but also from a lack of fidelity in capturing the interactions of the many complex subsystems that are typical of robotics (e.g., asynchronous message passing between sensors and processors, inaccuracies in the execution of actuation commands, uncertain state estimations, and the update frequency of the decision-making loop itself).
Some of the latter sources of simulation error can be reduced through software-in-the-loop (SIL) simulations that implement the entire autonomy software stack as well as the dynamics of the robot.
However, these autonomy stacks typically cannot run faster than real time, which precludes their use for training RL policies.

We accordingly propose a three-layer simulation pipeline.
Initially a \textit{low-fidelity} simultation---which implements only the robot's physical dynamics---is used to train and verify subsystem policies.
This low-fidelity simulation makes a number of simplifying assumptions, e.g., that the robot perfectly observes its own state.
Next, the resulting compositional RL policies are integrated into the software stack that will be deployed on the true robot, before being tested in a \textit{high-fidelity} SIL simulation that relaxes the assumptions made by the low-fidelity simulation, e.g., 
state observations are now given by odometry-based estimates of the robot's location and pose.
Finally, the policies are deployed on the target robotic system of interest.
At every level of fidelity in this simulation pipeline, the performance of the subtask policies are assessed and used to close the iterative compositional RL loop, i.e., to update the HLM.

In an experimental case study we apply the proposed framework to train and deploy a compositional RL system that pilots a \textit{Warthog} unmanned ground robot.
We implement the low-level simulation using the \textit{Unity} video game engine \cite{juliani2018unity}.
The simulation of the robot's environment is constructed from photogrammetry data of the target operating environment in the real world.
We develop software to integrate the trained RL policies with the \textit{Army Research Lab's} autonomy stack, named \textit{Phoenix}, and test the integration in a SIL simulation before successfully deploying the policies on the ground robot.

Through this case study we observe that the proposed framework not only facilitates the process of training an RL-based system to complete the robot's navigation task, but it also enables efficient adaptation to changes in the environment and it allows the user to isolate and address challenging portions of the sim-to-real transfer.
In one instance, 
a subtask was failing to satisfy its subtask specification when deployed in the real world. 
Instead of changing the overall learning objective and restarting the training process for the entire task, the compositional framework allowed us to isolate and retrain only the individual subtask policy responsible for the failure.
The end result was a successful compositional policy that consistently completes the robot's task.

\section{Preliminaries}
\label{sec:problem_statement}

We model the robot's control task using a partially observable Markov decision process (POMDP).
A POMDP is a tuple \(\pomdp\) \(= (\mdpStateSet,\) \(\mdpActionSet,\) \(\mdpTransition,\) \(\pomdpObservationSet,\) \(\pomdpObservationFunction)\) where \(\mdpStateSet\) is a set of states, \(\mdpActionSet\) is a set of actions, \(\mdpTransition : \mdpStateSet \times \mdpActionSet \times \mdpStateSet \to [0,1]\) is a transition probability function, \(\pomdpObservationSet\) is a set of possible observations, and \(\pomdpObservationFunction : \mdpStateSet \times \pomdpObservationSet \to [0,1]\) is an observation probability function. 

In general, a policy within the POMDP is a function \(\policy : (\pomdpObservationSet \times \mdpActionSet)^* \times \pomdpObservationSet \times \mdpActionSet \to [0,1]\) that maps histories of observations and actions \(\pomdpObservation_{0} \mdpAction_{0} \ldots \pomdpObservation_{t} \in (\pomdpObservationSet \times \mdpActionSet)^* \times \pomdpObservationSet \) to distributions over actions \(\mdpAction \in \mdpActionSet\).
Implementing policies over histories of arbitrary length is impractical.
Instead, it is common to define policies that map fixed-length histories of observations to distributions over actions.
We use RL algorithms to learn such policies \(\policy\) in \(\mdp\) \cite{sutton2018reinforcement,SpinningUp2018}.


\section{Compositional Reinforcement Learning}
\label{sec:methods}

Instead of taking a monolithic approach that searches for a single policy \(\policy\) accomplishing some objective in \(\pomdp\), we decompose the overall task of interest into subtasks and we define a \textit{high-level model} (HLM) to reason over \textit{compositions} of subtasks.
For the sake of completeness, below we briefly present definitions of tasks, subtasks, HLMs, and subtask compositions that are adapted from \cite{neary2022verifiable}.
However, for a more detailed discussion surrounding the motivation and intuition behind these definitions, we refer the reader to \cite{neary2022verifiable}.

\subsection{Tasks, Subtasks, and (Sub)Task Specifications}
\label{sec:subtasks}

We define a task in POMDP \(\pomdp\) as the tuple \((\mdpInitialState, \targetStateSet)\) that consists of an initial state \(\mdpInitialState \in \mdpStateSet\) and a target set of states \(\targetStateSet \subseteq \mdpStateSet\) that the robot must reach.
Furthermore, let \(\mathbb{P}^{\mdpInitialState}_{\mdp}(\Diamond_{} \targetStateSet| \policy)\) denote the probability of reaching the target set from \(\mdpInitialState\) under a particular policy \(\policy\).
We then define a \textit{task specification} as the requirement that \(\mathbb{P}^{\mdpInitialState}_{\mdp}(\Diamond_{} \targetStateSet| \policy) \geq 1- \hlmFailProb\) for some allowable probability of failure \(\hlmFailProb \in [0,1]\).

We similarly define a \textit{subtask} as a tuple \(\subtask = (\subtaskInitialStateSet_{\subtask}, \subtaskFinalStateSet_{\subtask})\), where \(\subtaskInitialStateSet_{\subtask} \subseteq \mdpStateSet\) is a set defining the subtask's \textit{entry conditions} and \(\subtaskFinalStateSet_{\subtask} \subseteq \mdpStateSet\) is a set defining the subtask's \textit{exit conditions}.
The objective of a \textit{subtask policy} \(\policy_{\subtask}\) is to reach an exit condition \(\mdpState' \in \subtaskFinalStateSet\) from any one of the subtask entry conditions \(\mdpState \in \subtaskInitialStateSet\).
We note that this definition is similar to the popular \textit{options} framework, defined by \cite{sutton1999between}.
A \textit{subtask specification} is then defined as the requirement that \(\mathbb{P}^{\mdpState}_{\mdp}(\Diamond_{} \subtaskFinalStateSet_{\subtask} | \policy_{\subtask}) \geq \bernoulliProb_{\subtask}\) for every \(\mdpState \in \subtaskInitialStateSet_{\subtask}\) and for some \(\bernoulliProb_{\subtask} \in [0,1]\).
In words, for subtask policy \(\policy_{\subtask}\) to satisfy the subtask specification, it must reach the target set with a minimum probability of \(\bernoulliProb_{\subtask}\) from any state in the subtask's set of entry conditions.

We define a collection \(\subtaskSet = \{\subtask_{1}, \subtask_{2}, ..., \subtask_{\numSubtasks}\}\) of subtasks to be \textit{composable}, if and only if for every \(i,j \in \{1,2,\ldots, \numSubtasks\}\), either \(\subtaskFinalStateSet_{\subtask_i} \subseteq \subtaskInitialStateSet_{\subtask_j}\) or \(\subtaskFinalStateSet_{\subtask_i} \cap \subtaskInitialStateSet_{\subtask_j} = \emptyset\).
Furthermore, we define a collection \(\subtaskSet\) of subtasks to be \textit{compatible} with an overall task \((\mdpInitialState, \targetStateSet)\) if and only if the following three conditions hold: 1) there exists at least one \(i \in \{1,\ldots, \numSubtasks\}\) such that \(\mdpInitialState \in \subtaskInitialStateSet_{\subtask_{i}}\), 2) there exists at least one \(i \in \{1,\ldots,\numSubtasks\}\) such that \(\subtaskFinalStateSet_{\subtask_{i}} = \targetStateSet\), 3) for every \(i \in \{1,\ldots, \numSubtasks\}\) either \(\subtaskFinalStateSet_{\subtask_{i}} = \targetStateSet\) or \(\subtaskFinalStateSet_{\subtask_{i}} \cap \targetStateSet = \emptyset\).

\subsection{The High-Level Model and Compositions of Subtasks}
\label{sec:hlm}

Given a composable collection \(\subtaskSet = \{\subtask_{1}, \subtask_{2}, \ldots, \subtask_{\numSubtasks}\}\) of user-defined subtasks that are compatible with a task \((\mdpInitialState, \targetStateSet)\), we define a \textit{high-level state abstraction} as the equivalence relation \(\eqRelation \subseteq \mdpStateSet \times \mdpStateSet\).
Two states \(\mdpState\) and \(\mdpState'\) are related by \(\eqRelation\), denoted \((\mdpState, \mdpState') \in \eqRelation\), if and only if the following two conditions hold.
\begin{align*}
	\begin{array}{l}
		 \text{1. For every \(\controller \in \controllerSet, \mdpState \in \controllerInitialStateSet_{\controller}\) if and only if \(\mdpState' \in \controllerInitialStateSet_{\controller}\), and}, \\
		 \text{2. \(\mdpState \in \controllerFinalStateSet_{targ}\) if and only if \(\mdpState' \in \controllerFinalStateSet_{targ}\).}
	\end{array}
\end{align*}%

Given the definition of a high-level state abstraction, we now define the high-level model (HLM) as a parametric MDP (pMDP) \(\abstractMDP = (\abstractStateSet, \abstractInitialState, \abstractSuccessState, \abstractFailureState, \controllerSet, \abstractTransition)\).
Here, \(\abstractStateSet\) is a set of high-level states defined as the collection of all equivalence classes \([\mdpState]_{\eqRelation} = \{\mdpState' \in \mdpStateSet | (\mdpState, \mdpState') \in \eqRelation\}\) induced by \(\eqRelation\), \(\abstractInitialState \in \abstractStateSet\) is an initial high-level state defined as the equivalence class \([\mdpInitialState]_{\eqRelation}\) containing \(\mdpInitialState \in \mdpStateSet\), \(\abstractSuccessState \in \abstractStateSet\) is a high-level goal state defined as the equivalence class \([\mdpState]_{\eqRelation}\) such that \(\mdpState \in \targetStateSet\), and \(\abstractFailureState \in \abstractStateSet\) is an additional high-level state that is only transitioned to when the task is failed.
Meanwhile, \(\abstractTransition : \abstractStateSet \times \subtaskSet \times \abstractStateSet \to [0,1]\) is a parametric transition probability function defined as,
\begin{align*}
\abstractTransition(\abstractState, \controller, \abstractState') = \begin{cases} 
  \bernoulliProb_{\controller}, & if\; \; \controller \in \controllerSet(\abstractState), \;\; \abstractState' = succ(\controller) \\
  1 - \bernoulliProb_{\controller}, & if \; \; \controller\in \controllerSet(\abstractState), \; \; \abstractState' = \abstractFailureState\\
  0, & \text{otherwise,}
\end{cases}\end{align*}
where \(\controllerSet(\abstractState) = \{\subtask \in \subtaskSet | \mdpState \in \subtaskInitialStateSet_{\subtask} \textrm{ for all } \mdpState \in \abstractState\}\) denotes the set of \textit{available subtasks} from high-level state \(\abstractState\), \(succ(\subtask)\) denotes the unique high-level successor state of subtask \(\subtask\) (i.e. \(succ(\subtask) = [\mdpState]_{\eqRelation}\) such that \(\mdpState \in \subtaskFinalStateSet_{\subtask}\)), and \(\bernoulliProb_{\subtask} \in [0,1]\) are parameters associated with the separate subtasks \(\subtask \in \subtaskSet\).

Intuitively, the states of HLM \(\abstractMDP\) correspond to sets of states in POMDP \(\pomdp\) from which the same collection of subtasks may be initiated.
If a subtask \(\subtask\) is initiated from a given high-level state \(\abstractState\), then with probability \(\bernoulliProb_{\subtask}\) the HLM transitions to a successor state that corresponds to the successful completion of the subtask, and with probability \(1 - \bernoulliProb_{\subtask}\) it transitions to the failure state \(\abstractFailureState\).
We note that the parameter \(\bernoulliProb_{\controller}\) may thus be interpreted as the probability of subtask \(\subtask\) being successfully completed, given that it is initiated from a state within its set of entry conditions.

A composition of subtasks is defined by a \textit{meta-policy} \(\abstractPolicy : \abstractStateSet \times \subtaskSet \to [0,1]\) that maps high-level states to distributions over subtasks.
The execution of a composition of subtask policies thus proceeds as follows.
From initial state \(\mdpInitialState\), the meta-policy conditions on the corresponding high-level initial state \(\abstractInitialState\) to select a subtask \(\subtask\) to execute.
The corresponding subtask policy \(\policy_{\subtask}\) is executed in POMDP \(\pomdp\) until it reaches an exit condition \(\subtaskFinalStateSet_{\subtask}\), at which point it relinquishes control.
The meta-policy then uses the corresponding high-level state \(\abstractState = succ(\subtask)\) to select the next subtask to execute and the process continues.
If the subtask policy fails to reach an exit condition (e.g., it does not relinquish control within some user-defined maximum allowable time), then the task is considered failed and the HLM transitions to \(\abstractFailureState\).

Let \(\mathbb{P}_{\mdp}^{\mdpInitialState}(\Diamond \targetStateSet | \abstractPolicy, \policy_{\controller_{1}}, \ldots, \policy_{\controller_{\numControllers}})\) denote the probability that such a composition of subtask policies reaches the target set \(\targetStateSet\) from initial state \(\mdpInitialState\) within POMDP \(\pomdp\).
Given an allowable failure probability \(\hlmFailProb \in [0,1]\), our objective is to find a collection of subtask policies \(\{\policy_{\subtask_{1}}, \ldots, \policy_{\subtask_{\numSubtasks}}\}\) and a meta-policy \(\abstractPolicy\) such that the composition satisfies the task specification \(\mathbb{P}_{\mdp}^{\mdpInitialState}(\Diamond \targetStateSet | \abstractPolicy, \policy_{\controller_{1}}, \ldots, \policy_{\controller_{\numControllers}}) \geq 1-\hlmFailProb\).

\subsection{Automatic Decomposition of Task Specifications}
\label{sec:compositional_rl}

We note that the meta-policy may be viewed as a Markovian policy within the HLM \cite{puterman2014markov}.
We may thus define the probability \(\mathbb{P}_{\abstractMDP}^{\abstractInitialState}(\Diamond \abstractSuccessState | \abstractPolicy, \bernoulliProb_{\controller_{1}}, \ldots, \bernoulliProb_{\controller_{\numControllers}})\) of \(\abstractPolicy\) reaching the high-level goal state \(\abstractSuccessState\) from the high-level initial state \(\abstractInitialState\),1.0 given values for the transition parameters \(\bernoulliProb_{\subtask_{1}}, \ldots, \bernoulliProb_{\subtask_{\numSubtasks}}\).

We briefly recall the conclusion of Theorem 1 from \cite{neary2022verifiable}, which states that if every subtask policy \(\policy_{\subtask}\) satisfies the subtask specification defined by parameter \(\bernoulliProb_{\subtask}\), i.e., \(\mathbb{P}^{}_{\mdp}(\Diamond_{} \subtaskFinalStateSet_{\subtask} | \policy_{\subtask}, \mdpState) \geq \bernoulliProb_{\subtask}\) for every \(\mdpState \in \subtaskInitialStateSet_{\subtask}\), then 
\begin{equation*}
    \mathbb{P}_{\mdp}^{\mdpInitialState}(\Diamond \targetStateSet | \abstractPolicy, \policy_{\controller_{1}}, \ldots, \policy_{\controller_{\numControllers}})
    \geq
    \mathbb{P}_{\abstractMDP}^{\abstractInitialState}(\Diamond \abstractSuccessState | \abstractPolicy, \bernoulliProb_{\controller_{1}}, \ldots, \bernoulliProb_{\controller_{\numControllers}}).
\end{equation*}

So, if we pick values for parameters \(\bernoulliProb_{\subtask_{1}}, \ldots, \bernoulliProb_{\subtask_{\numSubtasks}}\) and simultaneously find a meta-policy \(\abstractPolicy\) such that \(\mathbb{P}_{\abstractMDP}^{\abstractInitialState}(\Diamond \abstractSuccessState | \abstractPolicy, \bernoulliProb_{\controller_{1}}, \ldots, \bernoulliProb_{\controller_{\numControllers}}) \geq 1-\hlmFailProb\), then we may conclude that the composition of subtask policies defined by \(\abstractPolicy\) will satisfy the task specification of interest.
A parameter synthesis problem that simultaneously solves for a meta-policy and for a collection of values \(\bernoulliProb_{\subtask_{1}}, \ldots, \bernoulliProb_{\subtask_{\numSubtasks}}\) that minimize \(\sum_{i=1}^{\numSubtasks} \bernoulliProb_{\subtask_{i}}\) while ensuring this condition holds may be formulated as a bilinear program.
We exclude a full statement of this optimization problem due to space constraints, however, a detailed description of the problem is available in \cite{neary2022verifiable}.

\section{The Multifidelity Sim-to-Real Pipeline}
\label{sec:sim_pipeline}

We integrate the above framework for compositional RL with a multifidelity simulation of a wheeled ground robot and its operating environment. 
This pipeline allows for efficient training and testing of subtask policies in a \textit{low-fidelity} simulation that can be run faster than real time.
However, it also allows for the nuanced interactions between the trained policies and the existing software stack to be tested in a \textit{high-fidelity} simulation before deployment.

To ground the paper's ideas in a concrete problem setting, we present constructions for the states, actions, observations, and rewards that are specific to wheeled robot navigation tasks.
However, we emphasize that the presented framework for compositional RL and for multifidelity simulation may easily be adapted to different robotic systems, or to handle different task objectives and decision-making considerations.
We also remark that while we develop the proposed framework with RL algorithms in mind, both the HLM and the multifidelity simulation pipeline may be used to develop and test any collection of subtask policies \(\policy_{\subtask}\), so long as they all satisfy their subtask specifications.
Indeed, the subtask policies could be synthesized using RL algorithms, algorithms for planning and control, or using any combination thereof.

\newcommand{\vel}{v}
\newcommand{\position}{x}
\newcommand{\orientation}{\omega}
\newcommand{\linearVel}{\vel_{x}}
\newcommand{\angularVel}{\vel_{\omega}}

\subsection{Modeling the Robotic System}
\label{sec:physical_system}
We use the POMDP \(\pomdp_{true}\) \(= (\mdpStateSet,\) \(\mdpActionSet,\) \(\mdpTransition_{true},\) \(\pomdpObservationSet,\) \(\pomdpObservationFunction_{true})\) to represent the true physical system of interest. 
The robot's state \(\mdpState\) describes its position \( \position \in \mathbb{R}^{3}\), velocity \(\vel_{\position} \in \mathbb{R}^{3}\), orientation \(\orientation \in \mathbb{H}\), and angular velocities \(\vel_{\orientation} \in \mathbb{R}^{3}\).
So, the set of all possible states is given by \(\mdpStateSet = \mathbb{R}^{9} \times \mathbb{H}\), where \(\mathbb{H}\) denotes the quaternions.
The robot's permissible actions correspond to linear and angular velocity commands, i.e., \(a = [\linearVel, \angularVel] \in \mdpActionSet\), where \(\linearVel\) corresponds to velocity along the forward-backward axis and \(\angularVel\) corresponds to angular velocity about the axis perpendicular to the ground plane.
The action set itself is given by a bounded rectangle in \(\mathbb{R}^{2}\), i.e., \(\mdpActionSet = [\linearVel^{min}, \angularVel^{min}] \times [\linearVel^{max}, \angularVel^{max}] \subseteq \mathbb{R}^{2}\), where \(\linearVel^{min}\), \(\linearVel^{max}\), \(\angularVel^{min}\), and \(\angularVel^{max}\) correspond to the minimum and maximum allowable linear and angular velocity commands.
The robot's observations \(\pomdpObservation \in \pomdpObservationSet\) contain information related to its current state, although these observations may be noisy and may additionally include other modes of information (e.g., lidar or camera data).
Meanwhile, the observation probability function \(\pomdpObservation \sim \pomdpObservationFunction_{true}(\cdot | \mdpState)\) is unknown.
The specific observation \(\pomdpObservation \in \pomdpObservationSet\) that is sampled will depend not only on the measurement noise in the robot's sensors, but also on the software that the robot implements to process these measurements and to estimate its state.
Similarly, the transition probability function \(\mdpState' \sim \mdpTransition_{true}(\cdot | \mdpState, \mdpAction)\) is unknown and depends on the physical dynamics of the robot, as well as on the software that implements the velocity commands (e.g., velocity commands are typically used to compute setpoints for the feedback controllers of the individual wheels). 

\subsection{The Low-Fidelity Dynamics-Only Simulator}
\label{sec:low_fidelity_sim}

The low-fidelity simulation models only the aspects of the system that are fundamental to the decision-making problem at hand, while making several simplifying assumptions to reduce its required compuational effort.
In particular, it simulates the physical dynamics of the ground robot, and the configuration of the robot's operating environment (e.g., the locations of the buildings and goal).
However, it assumes that sensor measurements and state estimations are perfectly accurate, and it does not implement the majority of the software stack required to operate the robot.

Conceptually, we use POMDP \(\pomdp_{low}\) \(= (\mdpStateSet,\) \(\mdpActionSet,\) \(\mdpTransition_{low},\) \(\pomdpObservationSet,\) \(\pomdpObservationFunction_{low})\) to model the low-fidelity simulation.
We note that the sets of states \(\mdpStateSet\), actions \(\mdpActionSet\), and observations \(\pomdpObservationSet\) are identical to those from \(\pomdp_{true}\): this is necessary to ensure that the inputs and outputs to the learned subtask policies \(\policy_{\subtask}\) are consistent across simulations.
By contrast, \(\mdpTransition_{low}\) and \(\pomdpObservationFunction_{low}\) will both be slightly different than their real-world counterparts.
In particular, state transitions \(\mdpState' \sim \mdpTransition_{low}(\cdot | \mdpState, \mdpAction)\) are governed by a physics simulator of the robot's dynamics, which is accurate enough to capture the robot's kinematics and the individual feedback controllers outputting motor torques, but which cannot perfectly represent more complex effects like motor nonlinearities or wheel slip.
Furthermore, to avoid the need to simulate algorithms for localization and mapping, we assume that \(\pomdpObservationFunction_{low}\) returns perfect observations of the robot's position and velocity.

\begin{figure*}[t]
    \centering
    \input{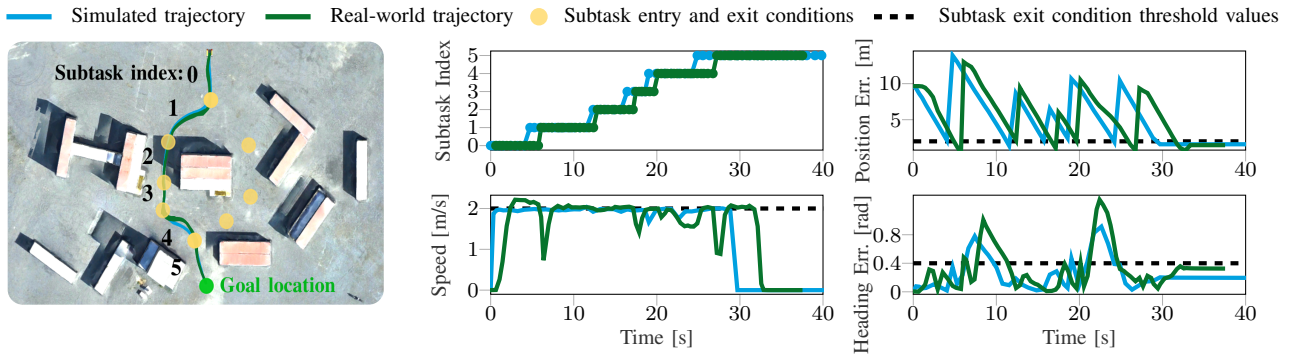}
    \caption{
        Left: Robot trajectories generated by a composition of subtask policies in the high-fidelity simulator (blue) and in the real world (green).
        The exit conditions of the possible subtasks are highlighted in yellow.
        Right: The index of the active subtask policy, the position and heading errors (w.r.t. the currently active subtask), and the robot's speed.
        The dashed black lines illustrate the threshold error and speed values that the robot must reach to successfully complete the active subtask. 
    }
    \label{fig:successful_compositional_policy}
\end{figure*}

\subsection{The High-Fidelity Software-in-the-Loop Simulator}
\label{sec:high_fidelity_sim}

The high-fidelity simulation builds on the low-fidelity dynamics simulation, but also includes the entire software stack that will be deployed to control the physical robot.
This software-in-the-loop (SIL) simulation thus relaxes the assumptions that the robot can perfectly observe its own state and that the velocity commands are instantaneously converted into motor-torque setpoints.
Instead, \(\pomdpObservation \sim \pomdpObservationFunction_{high}(\cdot | \mdpState)\) will be given by estimates of the robot's global position and velocity that are computed from local sensor measurements with simulated noise.
Meanwhile, the SIL simulation also captures the asynchronous message passing between the robot's sensors, post-processing scripts, subtask policies, and actuators. 
This affects the state transition probabilities \(\mdpState' \sim \mdpTransition_{high}(\cdot | \mdpState, \mdpAction)\) by changing the update rate of the decision-loop.
This asynchronicity between the RL-based policies and the rest of the simulation is closer to reality, but is markedly different from the implementation of the low-fidelity simulation, which sequentially alternates between sampling actions from the policy and advancing individual timesteps in the environment.

\subsection{Verifying the Subtask Policies and Refining the HLM}
\label{sec:refining_hlm}

Both the low-fidelity and high-fidelity simulations, as well as tests on physical hardware, are used to verify the performance of the compositional RL systems.
The results of these tests can be used to update the HLM and the definitions of the subtask entry and exit conditions, or to help manually troubleshoot issues with particularly challenging subtasks.
The result is an iterative procedure in which subtask policies are trained, their performance is tested in the multifidelity simulation pipeline, and the results of those tests are used to re-plan which subtasks should be used and trained further.

\section{An Experimental Case Study}
\label{sec:experiments}

To demonstrate the capabilities of the proposed framework, we apply it to train and verify compositional policies that control a \textit{Warthog} unmanned ground robot, illustrated in the bottom right of Figure \ref{fig:abstract_figure}.

\textit{The autonomous navigation task:}
We consider the navigation task of reaching the goal location (illustrated on the left in Figure \ref{fig:successful_compositional_policy}) with a probability of at least \(\hlmFailProb = 0.95\).
We define \(10\) possible subtasks, allowing the robot to either navigate left or right around the central building.
Each of these subtasks is defined by an initial location and heading, as well as a goal location and heading.
The subtask entry conditions \(\subtaskInitialStateSet_{\subtask}\) are then defined as the collection of states such that the robot is within 3.0 meters of its this initial location and within 0.5 radians of the initial heading.
Similarly, the exit conditions \(\subtaskFinalStateSet_{\subtask}\) are defined such that the robot is within 1.0 meter and 0.4 radians of the goal location and heading.

We define these initial (and goal) locations and headings to be overlapping, i.e., the goal location of one subtask will coincide with the initial location of another.
This overlapping structure defines a subtask graph (similar to that illustrated on the left in Figure \ref{fig:abstract_figure}), which we use to construct the HLM.

\textit{Training the subtask policies:}
We train each of the possible subtask policies \(\policy_{\subtask}\) for one million training steps in the low-fidelity simulator.
We parametrize each policy, and its corresponding value function, as a neural network with two fully-connected hidden layers of 64 units with \(\textrm{tanh}(\cdot)\) activation functions.
We train these networks using the \textit{Stable-Baselines3} implementation of proximal policy optimization (PPO) algorithm with the default hyperparameters \cite{schulman2017proximal,stable-baselines3}.

Each training episode for subtask \(\subtask\) begins by sampling an initial state from the subtask's entry conditions \(\mdpState \in \subtaskInitialStateSet_{\subtask}\).
The episode terminates with a reward of \(+5\) when the robot reaches an exit condition \(\mdpState \in \subtaskFinalStateSet_{\subtask}\), and it terminates with a reward of \(-20\) whenever the robot collides with an obstacle.
At every other timestep, the robot is assigned a negative reward that is a linear combination of: its distance to the subtask goal, the magnitude of the difference between its own heading and the goal heading, and its change in heading since the last timestep.

\begin{table}[t]
\centering
 \begin{tabular}{|c | c | c | c | c | c | c |} 
 \hline
 Subtask & 
 \(\subtask_{0}\) & 
 \(\subtask_{1}\) & 
 \(\subtask_{2}\) & 
 \(\subtask_{3}\) & 
 \(\subtask_{4}\) & 
 \(\subtask_{5}\) \\
 \hline\hline
 \(\hat{\bernoulliProb}_{\subtask}\)
 & 1.00 
 & 0.98 
 & 1.00 
 & 1.00 
 & \cellcolor{black!20!white}0.90 
 & 0.97\\ 
 \hline
 \(\bernoulliProb_{\subtask}\)
 & 1.00 
 & 0.98 
 & 1.00 
 & 1.00 
 & \cellcolor{black!20!white}0.95 
 & 0.97\\
 \hline
\end{tabular}
\caption{
Top: Empirical estimates of the probability of subtask success.
Bottom: Automatically decomposed subtask specification values for the meta-policy in Figure \ref{fig:successful_compositional_policy}. 
}
\label{tab:sub_task_specifications}
\end{table}

\textit{Constructing the low-fidelity simulation:}
We train the subtask policies in the low-fidelity simulation, which we implement using the video game engine \textit{Unity} \cite{juliani2018unity}.
The robot dynamics are modeled using Unity's built-in physics engine, the robot's kinematics, and simulated PID controllers for each of its four wheels. 
Meanwhile, the simulation of the robot's environment is constructed from photogrammetry data of the real-world test site.
At each timestep, the robot observes: its relative position and orientation with respect to the end goal defined by its subtask, its heading relative to the location of this goal, and its linear and angular velocities. 
The minimum and maximum allowable linear and angular velocity commands, which define the boundaries of the robot's action set, are set to \( \linearVel^{min} = 0 [\frac{m}{s}]\), \(\linearVel^{max} = 2 [\frac{m}{s}]\), \(\angularVel^{min} = -1.0 [\frac{rad}{s}]\), and \(\angularVel^{max} = 1.0 [\frac{rad}{s}]\).

\textit{Verifying the compositional RL systems in simulation:}
After training each subtask policy \(\policy_{\subtask}\), we estimate its probability \(\hat{\bernoulliProb}_{\subtask}\) of subtask success by rolling out the learned policy \(100\) times in the low-fidelity simulator from initial states that are sampled uniformly from its entry conditions \(\controllerInitialStateSet_{\controller}\).

To decompose the task specification (complete the task with a probability of at least \(\hlmFailProb=0.95\)) into subtask specifications, we solve the HLM parameter synthesis problem discussed in \S \ref{sec:compositional_rl} using \textit{Gurobi}~\cite{gurobi}.
Recall that the output of this problem is a meta-policy \(\abstractPolicy\) and a collection of subtask specifications: acceptable values \(\bernoulliProb_{\subtask}\) of the lower-bounds on the probabilities of subtask success. 

By comparing the empirical estimates, \(\hat{\bernoulliProb}_{\subtask}\), to the subtask specification values \(\bernoulliProb_{\subtask}\), the framework automatically determine which of the subtask policies are underperforming with respect to their requirements.
These specific subtask policies are then trained further until they either satisfy their subtask specifications, or a pre-defined maximum training budget is exhausted.
In the latter case, the framework will automatically add the constraint \(\bernoulliProb_{\subtask} \leq \hat{\bernoulliProb}_{\subtask}\) to the parameter synthesis problem and re-solve it: alternate subtasks need to be selected by the meta-policy to complete the overall task.

Table \ref{tab:sub_task_specifications} illustrates the values of both \(\hat{\bernoulliProb}_{\subtask}\) and \(\bernoulliProb_{\subtask}\) for subtasks \(\subtask_{0}\) to \(\subtask_{5}\) in the example from Figure \ref{fig:successful_compositional_policy}.
In this case, subtask \(\subtask_{4}\) is the only subtask for which \(\hat{\bernoulliProb}_{\subtask_{4}} \leq \bernoulliProb_{\subtask_{4}}\).
After further training, however, we empirically observe that \(\hat{\bernoulliProb}_{\subtask_{4}} = 1.00\).
At this point all subtask policies satisfy their subtask specification and so we conclude that the compositional policy defined by meta-policy \(\abstractPolicy\) will satisfy the overall task specification.

\textit{Testing the compositional RL systems using the high-fidelity simulation:}
Before deploying the compositional subtask policies on the physical hardware, we test their integration with the existing ROS-based software stack.
In the high-fidelity simulation, the robot's observations are derived from its odometry software.
These observations are processed into an input format compatible with the subtask policy networks.
The outputs of the policy networks are in turn converted into, and published as, ROS \textit{Twist} messages.
We remark that the inference time required to evaluate the policy networks is minimal: even on the robot's onboard computers, the policy can be evaluated at \(80 Hz\).

\textit{Compositional RL systems trained in simulation lead to successful task completion on hardware:}
Figure \ref{fig:successful_compositional_policy} illustrates the result of deploying the trained compositional RL systems both in the high-fidelity simulator as well as on the robot hardware. 
The resulting trajectories are very consistent: the left figure illustrates four separate runs of the compositional policies on hardware and five separate runs in simulation.
These separate trajectories can hardly be told apart.
Meanwhile, the plots on the right of the figure illustrate a representative trajectory from both the simulation and from the experiments on hardware.
We observe that the high-fidelity simulator is very accurate in predicting the outcomes of this particular composition of subtask policies.

\begin{figure}
    \centering
    \input{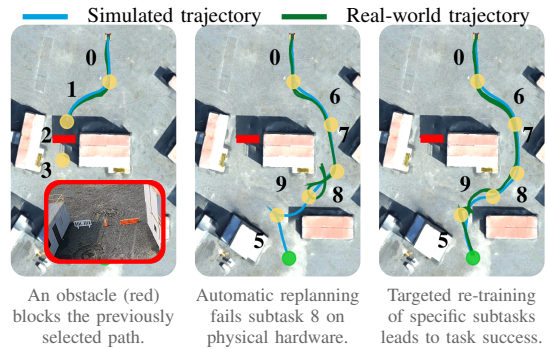}
    \caption{The compositional framework enables efficient adaptation to changes in the operating environment. 
    It also simplifies the process of resolving sim-to-real errors.
    }
    \label{fig:policy_adaptation}
\end{figure}

\textit{The framework automatically and efficiently adapts to environment changes:}
We now move a barricade into the robot's path, as illustrated on the left in Figure \ref{fig:policy_adaptation}.
We mirror this real-world environmental change by also including the obstacle in the low-fidelity simulator and by re-estimating \(\hat{\bernoulliProb}_{\subtask}\), the probabilities of subtask success. 
In this case, \(\hat{\bernoulliProb}_{c_{2}}\) has fallen to \(0.0\), and this value does not improve within the maximum allowed training budget for \(\policy_{\subtask_{2}}\). 
So, the meta-policy illustrated in Figure \ref{fig:successful_compositional_policy} is no longer guaranteed to satisfy the task specification in the low-fidelity simulation.

As described above, the framework thus adds the constraint \(\hat{\bernoulliProb}_{\subtask_{2}} \leq 0.0\) to the HLM parameter synthesis problem and re-solves it to obtain a new meta-policy and new subtask specifications.
This new meta-policy now selects subtasks that navigate past the other side of the central building.

Instead of restarting the training process for the entire compositional RL system, the framework re-uses policies \(\policy_{\subtask_{0}}\) and \(\policy_{\subtask_{5}}\). It only trains the newly required subtask policies \(\policy_{\subtask_{6}}\), \(\policy_{\subtask_{7}}\), \(\policy_{\subtask_{8}}\), and \(\policy_{\subtask_{9}}\) until they each satisfy their respective subtask specifications. 
The resulting composition of subtask policies successfully completes its task in 5 consecutive trials within the high-fidelity simulation, illustrated in blue in the middle of Figure \ref{fig:policy_adaptation}.

\textit{The framework simplifies the process of resolving sim-to-real errors:}
However, we observe from this same middle figure that while the composition of subtask policies is successful in simulation, it fails to satisfy the task when deployed in the real world.
In particular, subtask policy \(\policy_{\subtask_{8}}\) is never able to reach the heading angle necessary to complete its subtask.
This discrepancy between simulation and reality is likely due to errors in the simulated dynamics, particularly when the robot is attempting to turn on loose gravel.

However, this sim-to-real issue only causes a challenge for subtask \(\subtask_{8}\).
We accordingly adjust the heading angles that are used to define its entry \(\subtaskInitialStateSet_{\subtask_{8}}\) and exit \(\subtaskFinalStateSet_{\subtask_{8}}\) conditions, in order to reduce sharpness and precision of the turns that the robot is required to make in this area.
We then re-train only subtask policies \(\policy_{\subtask_{7}}\), \(\policy_{\subtask_{8}}\), and \(\policy_{\subtask_{9}}\) (note that by re-defining \(\subtaskInitialStateSet_{\subtask_{8}}\) and \(\subtaskFinalStateSet_{\subtask_{8}}\) we have changed the definitions of \(\subtaskFinalStateSet_{\subtask_{7}}\) and \(\subtaskInitialStateSet_{\subtask_{9}}\) as well).

The image on the right of Figure \ref{fig:policy_adaptation} illustrates the robot trajectories that result from this updated composition of subtask policies. 
By re-using the successful subtask policies and re-training only those causing challenges, we are able to efficiently adapt them until their composition consistently completes its task on hardware.

\section{Conclusions}
\label{sec:conclusions}

We propose a framework for compositional reinforcement learning (RL) within a multifidelity sim-to-real pipeline in order to facilitate the process of reliably deploying RL-based controllers on robot hardware. 
We demonstrate the framework's capabilities on an unmanned wheeled ground robot.
Future work will study how the framework might be used to train and deploy compositional multi-robot systems and vision-based subtask policies, as well as how the framework might be used to decompose and solve complex robotic tasks that are specified using, for example, temporal logics.

\section*{Acknowledgments}

This material is based upon work supported by, or in part by, the U. S. Army Research Laboratory and the U. S. Army Research Office under cooperative agreement numbers ARL W911NF-20-2-0132, ARL W911NF-19-2-0285, and ARO W911NF2010140. 
The views and conclusions contained in this document are those of the authors and should not be interpreted as representing the official policies, either expressed or implied, of the U.S. Army Research Laboratory, the U.S. Army Research Office, or the U.S. Government. 
The U.S. Government is authorized to reproduce and distribute reprints for Government purposes notwithstanding any copyright notation herein.

\bibliographystyle{IEEEtran}
\bibliography{bibliography}

\end{document}